\documentclass{llncs}
\usepackage{mdwlist}
\begin{document}

\belowcaptionskip 4mm

\title{Using Signals to Improve Automatic Classification of Temporal Relations}
\author{Leon Derczynski \and Robert Gaizauskas}
\institute{University of Sheffield, Regent Court, Sheffield S1 4DP, UK}

\maketitle

\begin{abstract}
Temporal information conveyed by language describes how the world around us changes through time. Events, durations and times are all temporal elements that can be viewed as intervals. These intervals are sometimes temporally related in text. Automatically determining the nature of such relations is a complex and unsolved problem. Some words can act as ``signals" which suggest a temporal ordering between intervals. In this paper, we use these signal words to improve the accuracy of a recent approach to classification of temporal links.
\end{abstract}

\section{Introduction}
The ability to order events, and the ability to determine which information is valid at a given time, are important in practical NLP. Effective automation of tasks such as summarisation and question answering require information extraction methods that can interpret information about time stored in documents. 

One difficult problem in temporal information extraction is the ordering of events. Although accurate event ordering has been the topic of much research~\cite{ahn2005towards,mani2007three,chambers2008jointly,llorens2010tipsem}, work using the temporal signals present in text -- for example, phrases such as \emph{after}, \emph{for the duration of} and \emph{while} -- has been limited, and often only yields a minimal benefit~\cite{min2007lcc}. Clearly these words contain temporal ordering information that human readers can access. This paper investigates the augmentation of a recent, high-performance temporal link classifier with information about temporal signals.

Our hypothesis is that signals provide information useful to TLINK classification. We also present data on signal usage within a temporally annotated corpus, in an attempt to gauge the likelihood of their being helpful and establish an upper bound on performance. After replicating existing work as a basis for comparison, we add signal-specific features and show how they lead to an improvement in classifier performance.

In this paper, we begin by describing the temporal annotation schema we have chosen to use (TimeML ~\cite{pustejovsky2004specification}) and provide a definition of temporal signals in the context of this paper (Sect.~\ref{background}. In Sect.~\ref{method}, we describe firstly how results from a previous experiment by Mani~et~al.~\cite{mani2007three} are replicated, and then detail the introduction of signal information into our system. Following this in Sect.~\ref{results} we detail our results, provide analysis in Sect.~\ref{analysis}, and conclude in Sect.~\ref{conclusion}.

\section{Background}
\label{background}
Here we will introduce the annotation used in this work, introduce problems with temporal signals, and cover some of the relevant literature.

\subsection{Temporal Annotation}
In order to capture temporal information well, a sophisticated annotation schema is required. We use the TimeML schema~\cite{pustejovsky2004specification}, which includes tags for event and time expression annotation (\texttt{<EVENT>} and \texttt{<TIMEX3>} respectively), as well as temporal relations between intervals (\texttt{<TLINK>}) and signal phrases (\texttt{<SIGNAL>}). The two largest resources of TimeML annotated text are TimeBank~\cite{pustejovsky2003timebank} and the AQUAINT TimeML corpus\footnote{Available for download from \underline{http://timeml.org/site/timebank/timebank.html}.}, which we merge to form a corpus for this work.

\subsection{Temporal Links}
Temporal links (or \textbf{TLINK}s) describe a temporal relation between two intervals, each of which is either an event or a time expression. Allen~\cite{allen1983maintaining} describes a set of relation types in terms of the interval endpoints. As our work is based on TimeML-annotated data, we use the set of TimeML relations, which are similar to Allen's. Each temporal link can optionally reference a signal.

\subsection{Signals}

Signals in TimeML are used to indicate multiple occurrences of events (temporal quantification) and also to mark words that indicate the type of relation between two intervals. For event ordering we are only interested in this latter use of signals. ``A University Grammar of English"~\cite{quirk1973university} lists a subset of these words in Sect.~10.5, ``Time Relaters".

For example, in the sentence \emph{John smiled \underline{after} he ate}, the word \emph{after} specifies an event ordering. This example could be represented in TimeML as follows:

\small
\begin{verbatim}
John <EVENT id="e1"> smiled </EVENT> <SIGNAL id="s1"> after </SIGNAL> 
he <EVENT id="e2"> ate </EVENT> .
<TLINK id="l1" eventID="e1" relatedToEvent="e2"
  relType="AFTER" signalID="s1" />
\end{verbatim}
\normalsize

TimeML allows us to associate text that suggests an event ordering (a signal) with a TLINK. To avoid confusion, it is worthwhile clarifying our use of the term ``signal". We use \textbf{SIGNAL} in capitals for tags of this name in TimeML, and \textbf{signal/signal word/signal phrase} for a word or words in discourse that describe the temporal ordering of an event pair. Examples of the signals found in TimeBank are provided in Table~\ref{tab:signalNgramFreq}. It is important to note that not every occurrence of text such as \emph{after} is a temporal signal. What is not shown due to space constraints is that a temporal signal such as \emph{after} may be used by (for example) 39 TLINKs labelled \textsc{after}, 17 labelled \textsc{before}, and four labelled \textsc{includes}; the signal text alone does not infer a single interpretation.

\footnotesize
\begin{table}
\begin{center}
\begin{tabular}{ | l | r | r | c | }
\hline
\textbf{Phrase} & \textbf{Corpus freq.} & \textbf{Occurrences as signal} & \textbf{Likelihood of being signal} \\
\hline
 subsequently & 3 & 3 & 100\%  \\
 after & 72 & 67 & 93\%  \\
 's & 10 & 8 & 80\%  \\
 follows & 4 & 3 & 75\%  \\
 before & 33 & 23 & 70\%  \\
 until & 36 & 25 & 69\%  \\
 during & 19 & 13 & 68\%  \\
 as soon as & 3 & 2 & 67\%  \\
\hline
\end{tabular}
\end{center}
\caption{A sample of phrases most likely to be annotated as a signal when they occur in TimeBank, which occur more than once in the corpus. All corpus data in this paper was provided by the CAVaT command-line tool~\cite{derczynski2010analysing}.}
\label{tab:signalNgramFreq}
\end{table}
\normalsize

\subsection{Previous work}

When temporally ordering events, it is intuitively likely that signal information may be useful. The trend in previous automated TLINK classification work has not been to directly target signals as a primary source of ordering information, although other attributes of annotated TLINKs and EVENTs have been exploited as training features. For example, the best known automatic TimeML annotation tool (TARSQI~\cite{verhagen2008temporal}) performs no SIGNAL annotation. Lapata and Lascarides~\cite{lapata2004inferring} worked with signals, using a restricted reference list of signal tokens instead of drawing signal text from human-annotated data. This work was only on same-sentence temporal links. Their accuracy at temporal relation classification was 70.7\%. Bethard and Martin~\cite{bethard2007timelines} included some features that described signals, where the \texttt{compl-word} feature (the signal text) was the 8$^{th}$ strongest in their set of features for temporal relation classification. However, this work has a number of limitations. First, it only uses the signal word and a simple relation type suggestion as features. It is also restricted to verb-clause construction TLINKs. Finally, The classifier only has to choose from a set of three TLINK classes (before, overlap, after).

\section{Method}
\label{method}

To explore the question of whether signal information can be successfully exploited for TLINK classification, we proceed as follows. First we re-implement a well-known TLINK relation classifier with state-of-the-art accuracy. Then we add various signal-related features to the classifier to investigate their impact on classification performance. The approach we have replicated as closely as possible is from Mani et al.~\cite{mani2006machine}. In brief, the method was as follows.

Firstly, the set of possible relation types was reduced by applying a mapping. For example, as \texttt{a} \textsc{before} \texttt{b} and \texttt{b} \textsc{after} \texttt{a} describe the same ordering between events \texttt{a} and \texttt{b}, we can flip the argument order in any \textsc{after} relation to convert it to a \textsc{before} relation. This simplifies training data and provides more examples per temporal relation class. Secondly, the following information from each TLINK is used as features: event class, aspect modality, tense, negation, event string for each event, as well as two boolean features indicating whether both events have the same tense or same aspect. Thirdly, we trained and evaluated the predictive accuracy of the maximum entropy classifier from Carafe\footnote{Available at \underline{http://sourceforge.net/projects/carafe/}.} using 10-fold cross-validation.

\footnotesize
\begin{table}
\begin{center}
\begin{tabular}{| l | c | c c | c |}
\hline
\textbf{Corpus} & \textbf{Total TLINKs} & \multicolumn{2}{|c|}{\textbf{With SIGNAL}} & \textbf{Without SIGNAL} \\
\hline
TimeBank v1.2 & 6418 & 718 & (11.2\%) & 5700 \\
AQUAINT TimeML v1.0 & 5365 & 178 & (3.3\%) & 5187 \\
ATC (combined) & 11783 & 896 & (7.6\%) & 10887 \\
ATC event-event & 6234 & 319 & (5.1\%) & 5915 \\
\hline
\end{tabular}
\end{center}
\caption{TLINKs and signals in our data.}
\label{tab:atcSignalCounts}
\end{table}
\normalsize

TLINK data came from the union of TimeBank v1.2a and the AQUAINT TimeML corpora. As the corpus used in the previous work by Mani et al. (TimeBank v1.2a) is not publicly available, we used TimeBank v1.2. This use of a publicly-available version of TimeBank instead of a private custom version was the only change from the previous method. In this work we only examine event-event links, which make up 52.9\% of all TLINKs in our corpus (See Table~\ref{tab:atcSignalCounts}).

We will later (Sect.~\ref{signalClassification}) add features that require data to be separated into test and training sets, with more sophistication required than that available in Carafe's maximum entropy classifier; thus, as well as performing 10-fold cross-validation (\textbf{XV}), we also split all event-event TLINKs into a training set of 4156 instances and an evaluation set of 2078 instances. 

\footnotesize
\begin{table}
\begin{center}
\begin{tabular}{| l | c | c |}
\hline
& \textbf{Predictive accuracy} & \textbf{Baseline} \\
\hline
Mani et al. results & 61.79\% & 51.6\% \\
Replicated results with our tools (10-fold XV) & 60.32\% & 53.34\% \\
Replicated results with our tools (train/test) & 60.04\% & 53.34\% \\
\hline
\end{tabular}
\end{center}
\caption{Results from replicating one of MITRE's TLINK classification experiments.}
\label{tab:tlinkReplication}
\end{table}
\normalsize

\subsection{Replicating Previous Work}

Table~\ref{tab:tlinkReplication} shows results from replicating the previous experiment on event-event TLINKs. The baseline listed is the most-common-class in the training data. We achieved a similar score of 60.32\% accuracy compared to 61.79\% in the previous work. The differences may be attributed to the non-standard corpus that they use. The TLINK distribution over a merger of TimeBank v1.2 and the AQUAINT corpus differs from that listed in the paper.

\subsection{Introducing Signals to the Feature Set}
\label{signalClassification}

To add information about signals to our training instances, we use the extra features described below; the two arguments of a TLINK are represented by \textbf{e1} and \textbf{e2}.

\begin{itemize*}
\item \textbf{Signal phrase.} This shows the actual text that was marked up as a SIGNAL. From this, we can start to guess temporal orderings based on signal phrases. However, just using the phrase is insufficient. For example, the two sentences \emph{Run before sleeping} and \emph{Before sleeping, run} are temporally equivalent, in that they both specify two events in the order run-sleep, signalled by the same word \emph{before}.
\item \textbf{Textual order of e1/e2.}  The textual ordering of linked events can be reversed without affecting temporal order. Thus, it is important to know the textual order of events and their signals even when we know a temporal ordering. This feature assumes that the order event-signal-event is most prevalent in text; values are either e1-e2 or e2-e1.
\item \textbf{Textual order of signal and e1, signal and e2.}
These features describe the textual ordering of both TLINK arguments and a related signal. It will also help us see how the arguments of TLINKs that employ a particular signal tend to be textually distributed.
\item \textbf{Textual distance between e1/e2.} Sentence and token count between e1 and e2.
\item \textbf{Textual distance from e1/e2 to SIGNAL.} If we allow a signal to influence the classification of a TLINK, we need to be certain of its association with the link's events. Distances are measured in tokens.
\item \textbf{TLINK class given SIGNAL phrase.} Most likely TLINK classification in the training data given this signal phrase (or empty if the phrase has not been seen). Referred to as signal \textbf{hint}.Referred to as signal \textbf{hint}.
\end{itemize*}





\section{Results}
\label{results}

Moving to a feature set which adds SIGNAL information, including signal-event word order/distance data, 61.46\% predictive accuracy is reached. The increase is small when compared to 60.32\% accuracy without this information, but TLINKs that employ a SIGNAL in are a minority in our corpus (possibly due to under-annotation). It would be interesting to see the performance difference when classifying only TLINKs that use a SIGNAL.

There are in total 11783 TLINKs in the combined corpus, of which 7.6\% are annotated including a SIGNAL; for just TimeBank v1.2, the figure is higher at 11.2\% (see Table~\ref{tab:atcSignalCounts}). The proportion of signalled TLINKs in our data is lowest at 5.1\%.

\footnotesize
\begin{table}
\begin{center}
\begin{tabular}{| l | c | c |}
\hline
\textbf{Predictive accuracy} & XV & Split \\
\hline
Baseline (most common class) & 53.34\% & 53.34\% \\
Without signal features & 60.32\% & 60.04\% \\
With basic signal features & \textbf{61.46\%} & 60.81\% \\
With signal features including hint & n/a & \textbf{61.98\%} \\
\hline
\end{tabular}
\end{center}
\caption{TLINK classification with and without signal features, using both 10-fold cross validation and a one-third/two-thirds split between evaluation and training data.}
\label{tab:addSignalFeatures}
\end{table}
\normalsize

The results of extending the feature set over a split of signalled and un-signalled links is shown in Table~\ref{tab:signalFeaturesHelpful}, from a one-third/two-thirds evaluation/training split.
\footnotesize
\begin{table}
\begin{center}
\begin{tabular}{| l | c | c |}
\hline
\textbf{Predictive accuracy} & \textbf{Unsignalled links} & \textbf{Signalled links} \\
\hline
Baseline & 52.68\% & 64.21\% \\
Plain features & \textbf{62.05\%} & 55.65\% \\
Plain + signal features & 62.05\% & \textbf{69.57\%} \\
Plain + signal features + hint & 62.05\% & 41.72\%\\
\hline
\end{tabular}
\end{center}
\caption[Predictive accuracy at classification of signalled and non-signalled TLINKs]{Predictive accuracy from Carafe's maximum entropy classifier, using features that do or do not include signal information, over signalled and non-signalled TLINKs in ATC. The baseline is accuracy when the most-common-class is always assigned.}
\label{tab:signalFeaturesHelpful}
\end{table}
\normalsize

\section{Analysis}
\label{analysis}

From Table~\ref{tab:signalNgramFreq} we can estimate the probability that a word or word sequence can be annotated as a SIGNAL associated with a TLINK. This may be of use when annotating signals, especially in the AQUAINT TimeML corpus. In any case, given that our feature set might only be helpful to 5.1\% of event-event links in the ATC corpus (Table~\ref{tab:atcSignalCounts}), the maximum performance increase at predicting signalled links can be estimated.

Let us suppose that we have perfect signal discrimination and association. Suppose our extra features do not help TLINKs without SIGNALs, and that the increase in performance is due solely to better accuracy classification of TLINKs that use signals. Let accuracy at classifying this signalled minority be $a$. Given a proportion of signalled TLINKs $s$, and predictive accuracy of our classifier when using features that do not depend on signals $P_n$ (from Table~\ref{tab:addSignalFeatures}):

\begin{eqnarray}
P_n(1 - s) + as = 0.6146\\
\mathbf{a = 0.8381}
\end{eqnarray}

Thus, we may be classifying signalled TLINKs at over 80\% accuracy when using the augmented features. This indicates a significant increase in predictive accuracy for signalled event-event TLINKs from the previous accuracy of 60.32\%. This is a target for classification of signal-employing TLINKs.

It is hard to determine an external upper bound for the classification of signal-employing TLINKs because inter-annotator agreement (IAA) figures are only available for TimeBank, and not at this level of detail. However, we can see from~\cite{boguraev2007timebank} that TLINK IAA reached 0.55. One would have to refer to the original annotator data and identify those TLINKs which were marked as employing a signal to determine an IAA value just for TLINKs with an associated SIGNAL. IAA for signals was 0.77.

We have hypothesised that adding features to represent signals in TLINK classification will lead to an increase in predictive accuracy. To test this, we repeat the above experiment, which compared features includign and excluding signal information. Data was divided into TLINKs that employ a signal, and those that do not. We expected to see similar prediction accuracy from both feature sets when classifying TLINKs that do not use signals. The baseline was the most common class in the dataset.

If there is no performance difference between feature sets when classifying TLINKs that \emph{do} use signals our hypothesis is incorrect, or the features we used are bad representation. If signals are helpful, and our features capture information useful for temporal ordering, we expect a performance difference when evaluating signalled TLINKs. Results in Table~\ref{tab:signalFeaturesHelpful} support our hypothesis that signals are useful, but we are performing nowhere near the maximum level suggested above. Data sparsity is a problem here, as the combined corpus only contains 319 suitable TLINKs, and both source corpora evidence of signal under-annotation. The results also suggest that the signal hint feature was not helpful; this is the same result found in~\cite{bethard2007timelines}.

Exploring the strongest feature set (basic+signals; no hint), attempting to combat the data sparsity problem, we used 10-fold XV instead of a split; results are in Table~\ref{tab:tsfsxv}. This shows a distinct improvement in the predictive accuracy of signalled TLINKs using this feature set over the features in previous work.

\footnotesize
\begin{table}
\begin{center}
\begin{tabular}{ | l | c | c | c | }
\hline
\textbf{Predictive accuracy} & \textbf{Baseline} & \textbf{Plain features} & \textbf{Plain and signal features} \\
\hline
Unsignalled links & 52.68\% & \textbf{61.81\%} & 61.81\% \\
Only signalled links & 62.41\% & 60.32\% & \textbf{82.19\%} \\
\hline
\end{tabular}
\end{center}
\caption{TLINK predictive accuracy using 10-fold cross validation over signalled and non-signalled TLINKs}
\label{tab:tsfsxv}
\end{table}
\normalsize

\section{Conclusion}
\label{conclusion}
When learning to classify signalled TLINKs, there is a significant increase in predictive accuracy when features describing signals are used. This suggests that signals are useful when it comes to providing information for classifying temporal links, and also that the features we have used to describe them are effective.

Future work is focused on improving signal and TLINK annotation. We need to explore how to discriminate whether or not a string is used as a temporal signal in text. Next, after finding a temporal signal, we need to determine which intervals it temporally connects. Finally, we can attempt to annotate a temporal link based on the signal. Once finished, we can integrate all this into existing temporal annotation tools.

\bibliographystyle{plain}
\bibliography{signals}
\end{document}